# Which Review Aspect Has a Greater Impact on the Duration of Open Peer Review in Multiple Rounds? —— Evidence from Nature Communications

Haomin Zhou, Ruxue Han, Jiangtao Zhong, Chengzhi Zhang*
Nanjing University of Science and Technology, 200 Xiaolingwei Street, Nanjing, China
{zhouhm, hrx, zhong_jt, zhangcz}@njust.edu.cn

**Abstract**

**Purpose –** Peer review plays a crucial role in scientific writing and the publishing process, assessing the quality of research work. As the volume of paper submissions increases, peer review becomes increasingly burdensome, highlighting the importance of studying the duration of peer review. This study aims to explore the correlation between review aspect sentiment and the duration of peer review, as well as the differences in this relationship across different disciplines and review rounds. Thus helping authors make targeted revisions and optimizations to their papers while reducing the duration of peer review. Which enables authors' research findings to reach the academic community and public domain more rapidly.

**Design/methodology/approach -** The study employs a two-step approach to understand the impact of review aspects on the duration of peer review. First, it extracts fine-grained aspects from peer review comments and uses sentiment classification models to classify the sentiment of each review aspect. Then, it conducts a correlation analysis between review aspect sentiment and the duration of peer review. Additionally, the study calculates sentiment scores for various review rounds to explore the differences in the impact of review aspect sentiment on the duration of peer review across different review rounds.

**Findings -** The study found that there is a weak but significant negative correlation between the sentiment of review and the duration of peer review. Specifically, the aspect clusters such as Evaluation & Result and Impact & Research Value, exhibit a relatively stronger correlation with the duration of peer review. Additionally, the correlation between review aspect sentiments and the duration of peer review varies significantly in different review rounds.

**Originality/value -** The significance of this study lies in connecting peer review comments text with the peer review process. By analyzing the correlation between review aspects and the duration of peer review, it identifies aspects that have a greater impact on the duration of peer review. This helps improve the efficiency of peer review from the perspectives of authors, reviewers, and editors. Thus alleviating the burden of peer review and accelerating academic exchange and knowledge dissemination.

* Corresponding Author, E-mail: zhangcz@njust.edu.cn.

# 1. Introduction

Nowadays, with the rapid advancement of scientific research, there is an increasing production of academic papers. Simultaneously, one of the most effective methods for evaluating these papers is peer review (also known as refereeing). The system of peer review can be traced back to the publication of papers in the 18th century *British Medical Journal*. By the end of the 20th century, peer review had been adopted by the majority of biomedical journals as an academic assessment system (Benos et al., 2007). Peer review is the evaluation by field experts of the competence, significance, and originality of research findings within that field. It serves as a crucial measure in ensuring the quality of scientific information, reducing errors, and mitigating misinformation. It plays a vital role in the process of scientific writing and publication (Tennant, 2018; Kang et al., 2018).

From submission to final publication, academic papers often undergo multiple rounds of peer review. After receiving a manuscript from the submitting author, the editor/chair initiates the initial review to ensure that the paper meets the quality standards of the journal/conference. Papers that pass this stage proceed to the peer review phase. In the peer review phase, multiple reviewers typically evaluate the paper, providing their comments and recommendations (reject/minor revisions/major revisions/accept). The editor/chair then decides whether to proceed to the next round based on the reviewers' feedback. Authors are required to respond to and make improvements based on the feedback from each reviewer. After receiving the author's responses and the revised manuscript, the reviewers initiate the next review round. This process continues until all reviewers have made clear decisions. Finally, based on all the review comments and recommendations, the editor/chair makes the final decision for the paper. The time it takes for an academic article to go through the peer review process from submission to eventual acceptance is referred to as the duration of peer review.

Peer review comments are texts written by field experts to assess submitted papers in their field, which can be considered a special comment. Reviewers typically start with an overall evaluation of the paper and then provide detailed assessments of various aspects of the paper. It contains aspect sentiment information of reviewers, which can reflect the reviewer's focuses and sentiment tendencies.

In recent years, the volume of academic paper submissions is increasing rapidly, while the peer review process is becoming exceedingly lengthy, placing a heavy burden on the peer review system (Huisman & Smits, 2017). With an increasing number of academic journals opting to open up peer review data, there has been significant attention on how to improve the efficiency of peer review. Many scholars have focused on peer review, employing methods such as bibliometrics, sentiment analysis, and others to study peer review comments text (Ramachandran & Gehringer, 2011; Ghosal et al., 2019a; Ghosal et al., 2019b; Chakraborty et al., 2020; Vincent-Lamarre & Larivière, 2021; Day, 2022; Zhang et al., 2022; Han et al., 2023; Wu et al., 2024), peer review structure (Hua et al., 2019; Qin & Zhang, 2021; He et al., 2023; Waltman et al., 2023), peer review process (Bornmann & Daniel, 2010; Lee et al., 2013; Lyman, 2013; Mrowinski et al., 2016; Sabaj et al., 2016; Helmer et al., 2017; Varas Espinoza et al., 2020; Chong, 2021; Squazzoni et al., 2021; Hecht et al., 2021; Severin & Chataway, 2021; Segarra-Saavedra et al., 2023), and peer review application (Price & Flach, 2017; Bhatia et al., 2020; Wang et al., 2020; Bharti et al., 2023; Kousha & Thelwall, 2023; Thelwall & Kousha, 2023). These studies aim to enhance our understanding of, and potentially improve, the peer review process. They have made significant contributions to accurately assess the significance and value of peer review, enhancing its efficiency. However, few studies connected peer review comments text with the peer review process. Which aspects of reviews and revisions in peer review will have a greater impact on the duration of peer review? Investigating this issue is particularly important for improving peer review efficiency. Authors can focus on these aspects of reviews during the revision process to avoid repeated revisions. Editors/chairs can also make quick decisions based on the review comments related to these aspects. The sentiment of peer review comments may indicate the complexity or depth of revisions required. Negative feedback often signals the need for major revisions, potentially extending the review duration, while more positive feedback may correlate with a faster acceptance process. This paper extracts fine-grained aspects from peer review comments in *Nature Communications* and investigates the following questions:

***RQ1***: Are there differences in the relationships between the sentiment scores of review aspects and the duration of peer review across different disciplines?

***RQ2***: Are there differences in the correlation between the sentiment scores of review aspects in different review rounds and the duration of peer review?

Specifically, in *RQ1*, we explore which review aspects have a greater impact on the duration of peer review by analyzing the correlation between the sentiment scores of different review aspects and the duration of peer review across different disciplines and its differences. In *RQ2*, we further analyze the correlation between the sentiment scores of different review aspects in different review rounds and the duration of peer review to uncover variations in the aspects that reviewers focus on in different review rounds. Through the above efforts, this study investigates which review aspects have a greater impact on the duration of peer review in different review rounds, thereby contributing to the improvement of peer review efficiency from the perspectives of authors, reviewers, and editors/chairs.

# 2. Related work

This chapter provides a review and discussion of related work from two aspects. Firstly, it elucidates the research related to peer review mining, and then it discusses the research related to sentiment analysis within the peer review field.

## 2.1 An overview of research on peer review mining

With the opening of peer review corpora, scholars have begun to delve deeper into peer review comments, focusing on the specific content, structural information, and procedural details of reviews.

Peer review comment texts are documents authored by expert reviewers as part of the peer review process for submitted manuscripts. They encompass the assessment, summary, queries, and recommendations made by reviewers regarding the submitted manuscript. And contain a wealth of information and value. Hua et al. (2019) investigated the content and structure of peer review within the framework of argument mining by automatically detecting argumentative propositions proposed by reviewers and their types (such as evaluations or suggestions for improvements). They found that the usage of propositions varied in terms of quantity, type, and subject in peer review. Kang et al. (2018) found a high correlation between overall recommendations and verbal recommendations by analyzing the PeerRead online peer review dataset. Certain aspects of a paper, such as the presence of appendixes, may be related to a high acceptance rate. Fromm et al. (2021) proposed an argument mining-based approach to efficiently extract the most relevant parts of reviews, thereby assisting editors and reviewers in their decision-making process. Ni et al. (2021) based on papers published in *Nature Communications* and their peer review files, used a regression model to conclude that there was no significant difference in the number of citations between open peer-reviewed papers and non-open peer-reviewed papers. Meng et al. (2021) proposed a method for constructing multi-level aspects for peer review of academic articles. These multi-level aspects reflect key aspects of reviewers' focus and can help inexperienced authors optimize their research design and writing. Ghosal et al. (2022) manually annotated a peer review dataset from four dimensions, including the corresponding sections of the paper, the attributes of the review comments, the role of the review comments, and the importance of the review comments. Han et al. (2022), using *Nature Communications* as an example, constructed a structured multi-round peer review corpus. They analyzed the changes in the focus of different rounds of reviewers through aspect extraction, assisting authors in making targeted improvements to their papers. Qin & Zhang (2023) analyzed the distribution of peer review comment texts in the structural sections of research articles based on their position information. They found that the quantity of peer review comment texts in the materials, methods, and results sections was significantly higher than in the introduction and discussion sections. However, this distribution was not correlated with citation rates.

Various complex factors can influence the duration of peer review, including the subject of the article, the reviewers, the journal/conference, and the authors themselves. Liang et al. (2022) examined the relationship between the novelty of articles and their processing time based on journal paper indexed in PubMed. They found that articles with higher novelty tend to have longer processing times, and longer processing times are negatively associated with the impact of the manuscripts. Additionally, articles from reputable authors and institutions have the advantage of shorter processing times, although this advantage decreases with increasing article novelty. Sarigöl et al. (2017), using PLoS One as a case study, investigated whether co-authorship relationships between authors and journal editors affect the processing time of manuscripts. They discovered that the frequency of editors handling papers from previous co-authors was significantly higher than expected by random chance. This prior co-author relationship was significantly associated with faster manuscript processing times, reducing the average review processing time by 19 days. Björk & Solomon (2013) studied average publishing delays in papers sampled from the Scopus citation index. They found that the shortest overall delays occur in science technology and medical fields while the longest in social science, arts/humanities and business/economics. And most of the time difference between submission and acceptance was within individual journals, rather than between journals, disciplines, or journal size. Similarly, differences between specific journals explained most of the variation in the time between acceptance and publication. Sabaj et al. (2015) investigated how the time it takes for peer review and publication decisions is linked to reviewer agreement. They found that peer review duration is influenced by decision type and varies across disciplines based on reviewer agreement.

The overview of research on peer review mining is shown in Table 1.

Currently, scholars have conducted extensive research on the content, structure, and process of peer review comments, and they have also analyzed the relationship between peer review comments and citations. However, most previous studies have treated peer review comments as a whole, neglecting the characteristics of multiple rounds of peer review. And they have mostly used relatively coarse-grained review aspects, which cannot provide authors with targeted and specific recommendations for paper revision and optimization.

**Table 1 The overview of research on peer review mining**

| Author (Year) | Research Object | Discipline | Corpus Size | Data Source | Research Content | Main Findings |
|---|---|---|---|---|---|---|
| Hua et al. (2019) | Review comment | Deep Learning | 14,200 Reviews | ICLR, ACL, UAI et al. | Types of review comments. | The usage of propositions varied in terms of quantity, type, and subject in peer review. |
| Kang et al. (2018) | Review comment | Deep Learning, Computational Linguistics, Natural Language Processing | 10,770 Reviews | NIPS, ICLR, ACL, CoNLL et al. | Paper acceptance prediction based on review comments. | There was a high correlation between overall recommendations and verbal recommendations. |
| Ni et al. (2021) | Paper & Review comment | Bioscience, Health Science, Physics, Earth and Environmental Science, Sociology | 7,614 Papers & 2,293 Reviews | Nature Communications | Whether open peer review facilitate citations. | There was no significant difference in the number of citations between open peer-reviewed papers and non-open peer-reviewed papers. |
| Meng et al. (2021) | Review comment | Bioscience, Health Science, Physics, Earth and Environmental Science, Sociology | 187,971 Reviews | Nature Communications | Constructing multi-level aspects for peer review. | Reviewers primarily focused on paper’s results, impact, methods, table & figure. |
| Han et al. (2022) | Review comment | Health Science, Physics, Earth and Environmental Science, Sociology | 41,373 Reviews | Nature Communications | Characterizing review aspects in multiple rounds. | Reviewers firstly focuses on paper’s results, methods & impact. |
| Qin & Zhang (2023) | Paper & Review comment | Chemistry, Physics, Earth and Environmental Science | 7,279 Papers & Reviews | ACP | Distribution of review comments in the article's section structure. | The quantity of review comments in the materials, methods, and results sections was significantly higher than in the introduction and discussion sections. |
| Liang et al. (2022) | Paper | Biomedical Science, Life Science, Physics | 778,345 Papers | PLoS One, Nature, Vaccine et al. | Relationship between the novelty of articles and the duration of peer review. | Articles with higher novelty tend to have longer processing times, and longer processing times are negatively associated with the impact of articles. |
| Sarigöl et al. (2017) | Paper | Biomedical Science, Life Science | 82,742 Papers | PLoS One | Relationship between author-editor relations and the duration of peer review. | The prior co-author relationship between editors and authors was significantly associated with faster manuscript processing times. |
| Björk & Solomon (2013) | Paper | Arts/Humanities, Biomedicine, Business/Economics, Chemistry, Earth Science, Engineering, Mathematics, Physics, Social Science | 2,700 Papers | Scopus | The publication delays in scholarly peer-reviewed journals across disciplines, journal size and journal quality | Most of the time difference between submission and acceptance was within individual journals, rather than between journals, disciplines, or journal size. Similarly, differences between specific journals explained most of the variation in the time between acceptance and publication. |
| Sabaj et al. (2015) | Paper | Humanities, Engineering, University teaching | 369 Papers | Onomázein, Formación Universitaria, Información Tecnológica. | The relationship between time taken for peer review, publication decision, and level of agreement among reviewers. | The peer review duration is influenced by decision type and varies across disciplines based on reviewer agreement. |

## 2.2 An overview of research on sentiment analysis of peer review

According to the granularity of the text being analyzed, sentiment analysis research can be categorized into three levels: document-level, sentence-level, and aspect-level (Liu, 2015). Aspect-level sentiment analysis directly focuses on opinions and their objects, emphasizing the extraction and analysis of opinion objects. It can mainly be divided into two tasks: aspect extraction and aspect-level sentiment classification. Hu & Liu (2004) first proposed using the syntactic relationship between sentiment words and aspect words to extract product aspect words. Experiments showed that this method was effective and applicable in English language corpora. Subsequently, Qiu et al. (2011) introduced the Double Propagation (DP) algorithm, which leverages word dependencies to synchronously extract sentiment words and opinion objects (entities, aspects). Zeng et al. (2019) proposed a Local Context Focus (LCF) sentiment classification mechanism based on Multi-Head Self-Attention (MHSA). It utilizes Contextual Features Dynamic Masking (CDM) and Contextual Features Dynamic Weighting (CDW) layers to pay more attention to local context words. Additionally, LCF incorporates a shared BERT layer to capture internal long-term dependencies between local and global contexts. Experiments demonstrated that LCF-BERT achieved the highest accuracy at the time.

Some scholars have begun to apply sentiment analysis techniques to the field of peer review to extract aspect and sentiment information from peer review comments. Wang & Wan (2018) proposed a Multiple Instance Learning Network with Abstract Memory Mechanism (MILAM) to identify sentences with sentiment polarity from review comments. It can also identify sentences with opinion based on sentiment polarity scores. Ghosal et al. (2019b) manually annotated a peer review comment dataset from four layers, including the corresponding sections of the review comments in the paper, review comment aspects, the role of review comments, and the importance of review comments. They also labeled sentiment polarity for the first two layers to assess the sentiment of the reviewer towards different sections and aspects of the paper. Thelwall et al. (2020) introduced a sentiment analysis tool called PeerJudge, designed to detect praise and criticism in peer reviews. PeerJudge employs a lexicon-based sentiment analysis approach, utilizing a sentiment lexicon that includes initially manually coded sentiment words, as well as sentiment words that have been adjusted and added through machine learning techniques. Luo et al. (2021) studied the correlation between review sentiment and review scores. Their findings revealed a positive correlation between review sentiment and review scores, and they demonstrated that manually encoded review comments could accurately predict whether proposals would receive funding. Lin et al. (2023) introduced an OSA mechanism that incorporates PU-learning into sentiment analysis of peer review comments. They employed a two-step process from PU-learning to assign an opinion sentence weight to each sentence in peer review comments. This weight helps the sentiment analysis model pay more attention to opinion sentences, thereby improving the extraction of reviewer sentiment polarity.

The overview of research on sentiment analysis of peer review is shown in Table 2.

Currently, scholars have conducted extensive aspect sentiment analysis research in the field of peer review. However, these studies have not performed fine-grained aspect extraction of peer review comments. Instead, they often directly use the review aspects mentioned in the ACL conference review criteria, which are relatively coarse-grained and do not provide specific recommendations for authors to revise and improve their papers.

Table 2 The overview of research on sentiment analysis of peer review

| Author (Year) | Discipline | Corpus Size | Data Source | Research Content | Main Findings |
|---|---|---|---|---|---|
| Wang & Wan (2018) | Deep Learning | 4,392 Reviews | ICLR | Multiple Instance Learning Network with Abstract Memory Mechanism. | The overall consistency between peer review comments and recommendation decisions is good, except for borderline comments. |
| Ghosal et al. (2019) | Deep Learning | 1,199 Reviews | ICLR | Construct a peer review comment dataset. | Sentiment can help improve the effectiveness and transparency of peer review systems. |
| Thelwall et al. (2020) | Biomedical Science, Deep Learning | 7,333 Reviews | F1000、ICLR | Lexicon-based sentiment analysis tool PeerJudge. | The absence of negative review comments is a reliable indicator for approval, but the presence of moderate negative review comments may lead to approval or conditional approval decisions. |
| Luo et al. (2021) | Biomedical Science, Life Science, Materials and Chemistry, Sociology, Engineering Science | 652 Reviews | SFI、SNSF | The correlation between review sentiment and review scores. | There is a positive correlation between review sentiment and review scores, and manually encoded review comments could accurately predict whether proposals would receive funding. |
| Lin et al. (2023) | Deep Learning | 4,391 Reviews | ICLR | Introducing PU-learning into sentiment analysis of peer review comments. | The acceptance rate metric can be used to effectively predict the decision status of a paper. |

In summary, peer review comments contain a wealth of information that can be mined. Extracting aspects from review comments can help capture the opinions and focal points of reviewers, and further classifying sentiments regarding these aspects can reveal the sentiment attitudes of reviewers. Past research has often employed relatively coarse-grained aspects, mainly drawing from the aspects provided in the ACL conference guidelines. Additionally, previous studies have paid limited attention to the information about the review rounds in peer review, and there has been a lack of emphasis on connecting peer review comments with peer review process. This study aims to extract fine-grained aspects from peer review comments, employing sentiment classification models to automatically classify the review aspect sentiment. Following this, the study conducts a correlation analysis between review aspect sentiment and the duration of peer review to investigate which review aspects may exert a greater impact on the duration of peer review. Moreover, the research includes the computation of sentiment scores for different review rounds, to investigate the differences in the impact of review aspect sentiment on the duration of peer review across different review rounds.

The work most relevant to this paper is Liang et al. (2022), where they explored the relationship between the novelty of articles and their processing time. Similar to this paper, both studies investigate the correlation between aspects of articles and their the duration of peer review. They quantified the novelty of articles by calculating a disruption index based on the citation patterns of the articles in the PubMed-indexed journals. In contrast, this paper quantifies the review aspect sentiment through sentiment analysis of peer review comments.

# 3. Methodology

This paper starts from peer review comments data and utilizes peer review files of papers published in Nature Communications[1] between 2016 and 2020 as the source of data. Firstly, the multi-round peer review comments are divided into rounds to create a structured peer review corpus for Nature Communications. Subsequently, a review aspect extraction is performed, and the extracted aspect words are then subjected to aspect clustering to obtain review aspect clusters. Following this, a random selection of review sentences from the Nature Communications peer review corpus that contain review aspects is made for manual annotation of aspect sentiments. The annotated review aspect sentiment dataset is then divided into a training set and a test set. Various sentiment classification models are employed for training and testing, and the best model is obtained. Subsequently, the best model is used for aspect sentiment classification task on all corpus. Based on this, sentiment scores for each aspect cluster and the overall review are calculated for each peer review file. Subsequently, separate correlation analysis is conducted for the sentiment scores of each aspect cluster and the overall review with the duration of peer review to explore which review aspects might have a greater impact on the duration of peer review. Additionally, control variables such as article length, the number of authors, and the number of subjects are introduced. And we comprehensively analyze the correlation between review aspect sentiment and the duration of peer review across different review rounds. As a result, the framework of this study is illustrated in Figure 1.

[1] https://www.nature.com/ncomms/, 2021/4/1

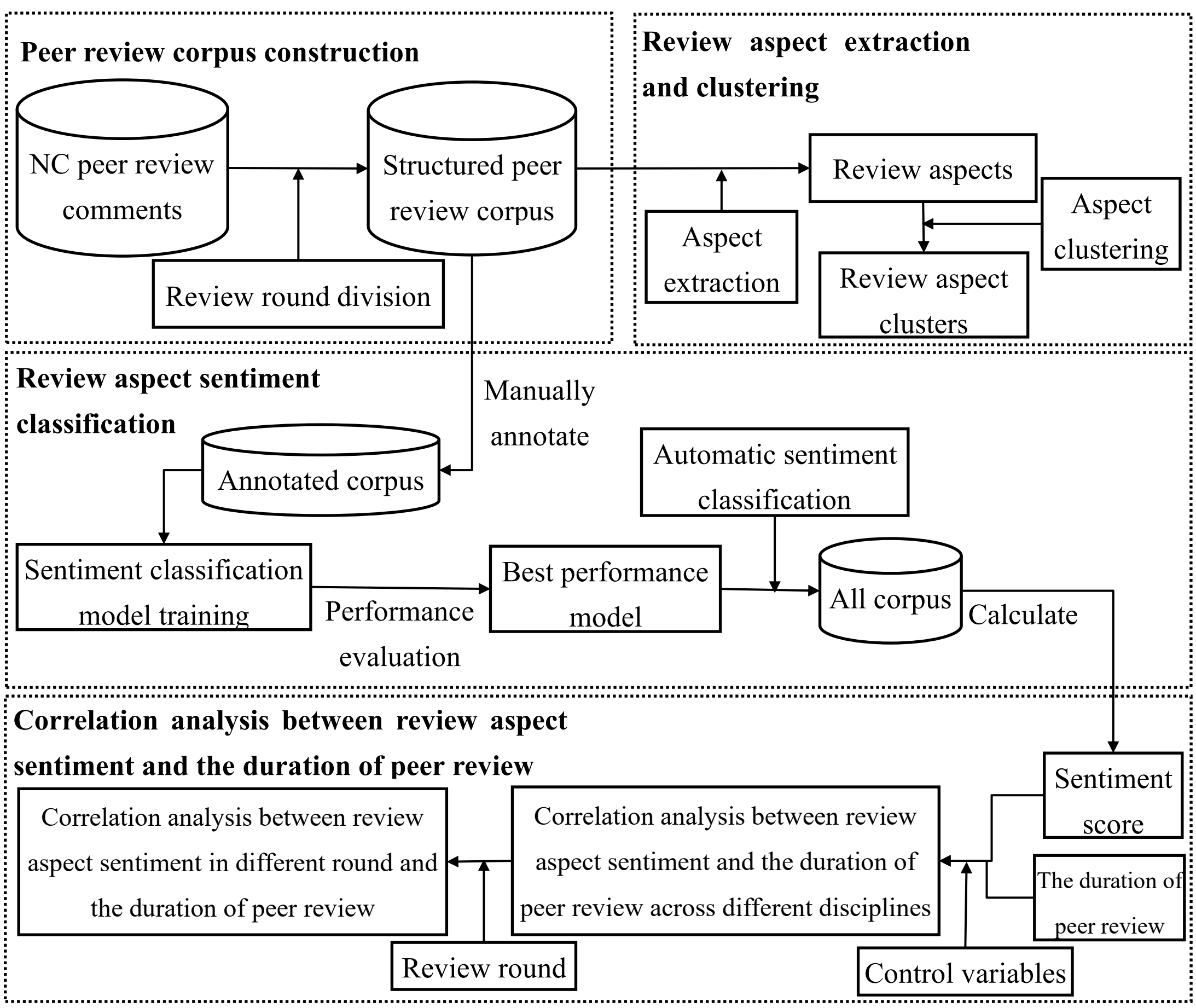

**Figure 1 The framework of this study**

## 3.1 Peer review corpus construction

This study utilizes the peer review corpus constructed by Han et al. (2022). They took *Nature Communications* as an example, employed text processing methods to build a multi-round peer review corpus, conducted review aspect extraction and clustering, performed sentiment classification at aspect level, and constructed a review aspect sentiment corpus. *Nature Communications* is an open-access journal that publishes high-quality research from various fields of the natural sciences. The papers published in this journal aim to present significant advancements that are relevant to experts in each respective field. The journal started opening the peer review reports of accepted papers in January 2016, with an open data rate of 60% in the first year. By 2019, this rate increased to 69%. The peer review process employed by the journal is a double-blind review.

They collected peer review comments on articles published in *Nature Communications* from 2016 to 2020, resulting in a total of 41,373 peer review comments. The data collection was conducted on April 1st, 2021. Initially, a web scraping tool was used to collect the peer review files in PDF format from the official website of *Nature Communications*. To facilitate subsequent machine processing and recognition, all the PDF files were uniformly converted into TXT format, which can be directly processed by computers[2]. Additionally, they gathered information about the duration of peer review, article authors,

[2] https://www.shipinzhuanhuan.cn/pdfconverter4/. We used Xunjie PDF Converter for PDF conversion. The accuracy of the conversion is very high, only a very small number of pdf in the conversion will become a mess.

disciplines, and article length.

Each peer review file contained one or more rounds of peer review comments, and all collected data were stored according to primary and secondary disciplines, as shown in Table 3.

Afterwards, they employed a combination of machine recognition and manual identification for the division of review round. The results of the division underwent manual verification to ensure accuracy. Finally, the segmented corpus was processed into sentences and stored in a database. An example of NC peer review corpus is illustrated in Table 4.

**Table 3 The statistics of NC peer review data**

| First-level disciplines | Number of second-level disciplines included | Number of peer review files |
|---|---|---|
| Biological sciences | 24 | 21683 |
| Earth and environmental sciences | 12 | 2588 |
| Health sciences | 17 | 6090 |
| Physical sciences | 9 | 10180 |
| Scientific community and society | 9 | 832 |
| Total | 71 | 41373 |

**Table 4 An example of NC peer review corpus**

| First-level disciplines | Second-level disciplines | Article ID | Review round | Review sentence number | Review sentence content |
|---|---|---|---|---|---|
| Biological sciences | biochemistry | s41467-017-00225-z | 1 | ##43 | *At least this approach should be better justified and perhaps it should be removed from the manuscript.* |
| Biological sciences | biochemistry | s41467-017-00225-z | 2 | ##23 | *Again, it would be nice if the corrected values of Wi in the present study were made more obvious.* |
| Biological sciences | biochemistry | s41467-017-00225-z | 3 | ##65 | *Are they allowing for some tiny flux of carbon going up the xylem or something?* |

And we further manually checked the conversion results to eliminate these errors.

## 3.2 Review aspect extraction and clustering

Han et al. used the Double Propagation (DP) algorithm to extract candidate aspect words from peer review comment texts, and after manually filtering out non-aspect words, they obtained a set of review aspects. The core idea of the DP algorithm is based on a bootstrapping strategy, requiring only a small number of seed sentiment words as input and no initial aspect words. The algorithm then uses syntactic analysis to identify the dependency relationships between sentiment words and aspect words, extracting a large number of aspect words from known sentiment words. Simultaneously, more sentiment words are extracted through the reverse extraction of the identified aspect words. This bidirectional extraction of sentiment and aspect words is iteratively repeated until no new aspect or sentiment words appear, completing the aspect extraction process (Qiu et al., 2011).

Then, they used the Affinity Propagation (AP) clustering method to cluster the review aspect words, obtaining candidate review aspect clusters. The core idea of AP clustering is to consider all data points as potential cluster centers. The method involves connecting every pair of data points to form a network (similarity matrix), and then calculating the cluster centers for each sample through message passing along the edges of the network (Frey & Dueck, 2007). Afterward, manual removal of non-review aspect clusters and merging of review aspect clusters with similar meanings were performed. In the end, they summarized and obtained 11 review aspect clusters, as shown in Table 5.

**Table 5 Review aspect clusters**

| **Review aspect clusters (abbreviation)** | **Review aspect words** | **Sample sentences** |
|---|---|---|
| Problem Definition/Idea**(PDI)** | definition/concept/issue/question/ problem/view/assumption/direction/... | This is a very nice study that tackles an important scientific ***question***. |
| Data/Datasets**(DAT)** | data/dataset/information/material/ sample/database/signal/noise/... | However, I have additional concerns about other new ***data***. |
| Tables & Figures **(TNF)** | image/figure/table/picture/graph/plot/ diagram/label/panel/legend/... | The ***figure legend*** of figure(2a) needs rephrasing. |
| Methodology & Model**(MET)** | model/method/approach/methodology/ mechanism/structure/strategy/... | The ***model*** is perhaps not vital to the key points of the manuscript. |
| Experiments & Operations**(EXP)** | experiment/training/testing/processing/ control/setup/feedback/adaptation/... | One simple ***experiment*** that could be done is synthesizing metal particles that are more spherical in nature. |
| Conclusions & Discussions**(CON)** | conclusion/claim/discussion/discovery/ finding/explanation/demonstration/... | What kind of meaningful ***conclusions*** can be reached by the investigator? |
| Index/Parameters **(IND)** | parameter/threshold/baseline/standard/t ime/metric/proportion/level/... | Other unmentioned ***parameters*** were the default settings of the software. |
| Evaluation & Result**(RES)** | measurement/analysis/result/accuracy/ estimation/evaluation/calculation/... | This would provide a conclusive ***result***. |
| Structure & Language**(SNL)** | description/expression/language/clarity /statement/understanding/wording/... | They ***claim*** that the change in charging of the particles is responsible for a change in the distance between the twisted pair. |
| Impact & Research Value**(IMP)** | impact/performance/contribution/ influence/novelty/effect/meaning/... | The usefulness of this ***effect*** is questionable, especially in real biological systems |

| Comparison & Correlation **(CMP)** | comparison/contrast/correlation/association/difference/similarity/... | A positive control such as lipofectamine is required for ***comparison***. |
|---|---|---|

## 3.3 Review aspect sentiment classification

This paper randomly extracted over 5000 sentences containing review aspects from the structured peer review corpus. These sentences were then annotated with aspect-level sentiment polarity. Each annotated data provided the content of the review sentence, aspect word, sentiment word, syntactic dependency relationship, the DOI of the corresponding article, and the URL.

In this paper, sentiment polarity was categorized into positive, neutral, and negative sentiments. During the annotation process, 1 represented positive sentiment, 0 represented neutral sentiment, and -1 represented negative sentiment. Three undergraduate students majoring in Information Management and Information Systems, all with a background in NLP research and relevant annotation experience, participated in the annotation work. Each annotation data was labeled by two annotators, with each annotator labeling two-thirds of the total annotated data. After the initial annotation, we evaluated the inter-annotator agreement using the Kappa coefficient (McHugh, 2012). The calculation involved dividing the number of agreed annotations by the total number of annotations. The consistency results were 0.62, 0.68, and 0.62, falling within the 0.6-0.8 range, indicating effective agreement. For annotation data where the two annotators disagreed, a third annotator reviewed and finalized the annotation. Finally, we conducted a statistical analysis of the annotated results, totaling 5063 sentences of review aspect sentiment. Among them, there were 1498 sentences with positive sentiment, 1652 sentences with neutral sentiment, and 1913 sentences with negative sentiment.

This study divided the annotated aspect sentiment dataset into training, testing, and validation sets in an 8:1:1 ratio. Various sentiment classification models were trained and tested for their performance, and the optimal model was selected for the task of aspect sentiment classification on the entire corpus. The study employed specialized models commonly used for Aspect-Based Sentiment Analysis (ABSA). The main sentiment classification models used in this study are as follows:

**Aspect Term Attention with LSTM (ATAE-LSTM)**: The ATAE-LSTM model utilizes attention mechanisms to focus on key information in the text related to specific aspects, enabling accurate fine-grained sentiment classification (Wang et al., 2016).

**Memory Network Model (MEMNET)**: The MEMNET model employs an external memory matrix to capture crucial information in the text. It utilizes an attention mechanism to weight and aggregate this information, thereby achieving sentiment classification (Tang, et al., 2016).

**Interactive Attention Network Model (IAN)**: The IAN model utilizes an interactive attention mechanism to learn attention over both context and target simultaneously. It generates representations for the target and context separately, enabling it to effectively capture the interplay between them (Ma et al., 2017).

**Attention-over-attention neural networks Model (AOA)**: AOA is a neural network model based on multiple attentions. It jointly models aspects and sentences in a coherent way, explicitly capturing the interactions between aspects and context sentences (Huang et al., 2018).

**Content Attention-Based Aspect Sentiment Classification Model (CABASC)**: CABASC employs an attention mechanism to focus on context content relevant to specific aspects. It calculates weighted representations for both aspects and context through attention mechanisms (Liu et al., 2018).

**Attention Encoding Network Model (AEN_BERT)**: AEN_BERT combines the power of pre-

trained BERT model with attention-based encoders to model both context and target aspects (Song et al., 2019).

**BERT Model based on Sentence Pair Classification (BERT_SPC)**: BERT_SPC constructs the input sequence as "[CLS] + Global Context + [SEP] + Aspect Word + [SEP]". The "[SEP]" labels are used to separate the two sentences, and the output vector corresponding to the "[CLS]" token serves as the representation for the entire sentence, utilized for category prediction (Devlin et al., 2019).

**Local Context Focus Model (LCF_BERT)**: LCF_BERT incorporates the concept of Local Context Focus (LCF) and Semantic Relative Distance (SeRD) to determine whether context words belong to the local context of a specific aspect. This approach allows LCF_BERT to discard words that are not relevant to the aspect word (Zeng et al., 2019).

## 3.4 Variables of correlation analysis between review aspect sentiment and the duration of peer review

This section primarily introduces and provides a descriptive statistics of the main variables involved in the correlation analysis between review aspect sentiment and the duration of peer review. The variables analyzed include the sentiment scores for different review aspect clusters and the duration of peer review of peer reviews. Control variables encompass article length, the number of authors, and the number of subjects.

### (1) Calculation of sentiment scores

This study quantifies the review aspect sentiment by calculating sentiment scores of each review aspect cluster for each peer review. The formula for calculating the sentiment score is as follows:

$$S = \frac{N_p - N_{ng}}{N_p + N_{ng} + N_{nu}} \tag{1}$$

Where, $N_p$ represents the number of positive sentiments, $N_{ng}$ represents the number of negative sentiments, $N_{nu}$ represents the number of neutral sentiments.

### (2) Calculation of the duration of peer review

The paper goes through varying lengths of time from submission to acceptance during the review process, known as the duration of peer review. In this study, the submission date and acceptance date of papers were obtained from the HTML versions of the papers on the *Nature Communications* website. The time difference between these two dates is then calculated to quantitatively represent the duration of peer review.

### (3) Calculation of control variables

The factors influencing the duration of peer review are highly complex and can be associated with various aspects such as the subjects of paper, reviewers, journal/conference, and authors themselves. Since *Nature Communications* maintains confidentiality regarding reviewer information, control variables related to reviewers cannot be obtained. As this study's data is sourced from a single journal, factors related to journal/conference do not need consideration. To mitigate the potential impact of other confounding factors, this study utilizes article length, the number of authors, and the number of subjects as control variables for the correlation analysis. The article length is quantified by calculating the number of words in the paper. The number of authors represents the number of credited authors of the paper. The number of subjects represents the number of subject keywords of the paper.

# 4. Results

This section will present and analyze the experimental results. First, we will discuss the results of sentiment classification of peer review comments. Following that, we will delve into the correlation analysis between review aspect sentiment and the duration of peer review across different disciplines. Finally, we will explore the correlation analysis between review aspect sentiment in different review rounds and the duration of peer review.

## 4.1 Results of review aspect sentiment classification

### 4.1.1 Results of the performance evaluation of sentiment classification model

This study employs four model evaluation metrics—Accuracy, Macro-Precision, Macro-Recall, and Macro-F1—to assess the performance of the utilized sentiment classification models. The following are the calculation formulas for each metric:

$$Accuracy = \frac{\mathrm{n_{correct}}}{\mathrm{n_{total}}} \quad (2)$$

$$Macro_P = \frac{1}{n}\sum_{i=1}^{n} P_i \quad (3)$$

$$Macro_R = \frac{1}{n}\sum_{i=1}^{n} R_i \quad (4)$$

$$Macro_F_1 = \frac{2*Macro_P*Macro_R}{Macro_P+Macro_R} \quad (5)$$

Where, $n_{correct}$ is the number of correctly classified, $n_{total}$ is the number of total classified, $n$ is the number of categories.

Results of the performance evaluation of the employed sentiment classification models are presented in Table 6.

**Table 6 The performance of sentiment classification models**

| Model | Accuracy(%) | Macro-P(%) | Macro-R(%) | Macro-F1(%) |
|---|---|---|---|---|
| AOA | 66.73 | 68.25 | 66.37 | 66.93 |
| IAN | 67.72 | 68.33 | 67.45 | 67.42 |
| CABASC | 70.09 | 70.74 | 69.58 | 69.68 |
| ATAE_LSTM | 70.48 | 71.74 | 70.09 | 70.39 |
| MEMNET | 71.17 | 71.75 | 71.09 | 71.25 |
| AEN_BERT | 77.10 | 79.91 | 76.18 | 76.71 |
| BERT_SPC | 79.66 | 80.18 | 79.57 | 79.75 |
| LCF_BERT_CDW | 80.47 | 81.24 | 80.05 | 80.33 |
| LCF_BERT_CDM | **81.24** | **82.64** | **80.90** | **80.77** |

From Table 6, it can be observed that the BERT-based models outperform other deep learning-based models in aspect sentiment classification task. The LCF-BERT model, which utilizes the Contextual Dynamic Masking (CDM) method for extracting contextual features, demonstrates the best performance. It achieves an Accuracy of 81.24%, Macro-Precision of 82.64%, Macro-Recall of 80.90%, and Macro-F1 of 80.77%, all surpassing the values of all other models. This model enhances its ability to extract and learn contextual features from the BERT shared layer by attenuating relatively less semantically informative features and focusing more on the local contextual words of specific aspects. Consequently, it achieves superior sentiment classification performance. Therefore, we ultimately employ the LCF-BERT model with the Contextual Dynamic Masking (CDM) method for aspect sentiment classification task on all corpus.

#### 4.1.2 Results of sentiment distribution of peer review comments

This study conducted a statistical analysis of sentiment polarities distribution for each review aspect cluster on all corpus. The specific results are illustrated in Figure 2.

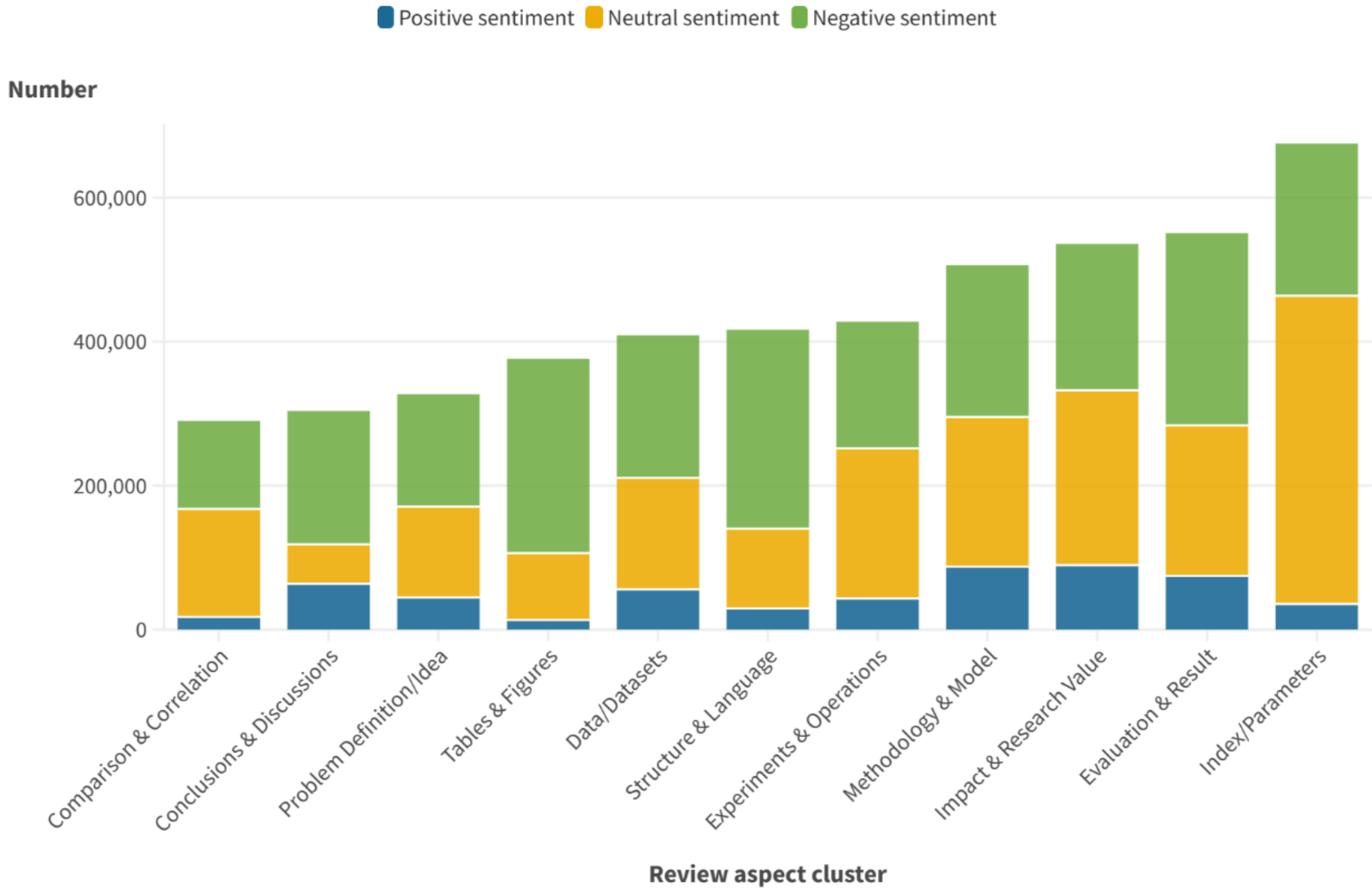

**Figure 2 Aspect sentiment distribution**

From Figure 2, it is evident that in peer review comments, the number of sentences containing neutral sentiment is the highest, followed by negative sentiment. Sentences containing positive sentiment are much less common, mainly because reviewers tend to focus on raising questions and providing critical feedback when evaluating papers, resulting in a lower occurrence of praising statements. Among all review aspect clusters, the highest proportion of negative sentiment are *Tables & Figures* and *Structure & Language*. This suggests that when reviewers focus on these aspects, they are more likely to provide critical feedback or suggestions, with fewer instances of praising statements. The aspect cluster *Index/Parameters* shows a significantly higher proportion of neutral sentiment compared to other aspect clusters, possibly because reviewers often provide an overview of the paper's content when discussing this aspect, with fewer sentiment expressions. Conversely, the aspect cluster *Conclusions & Discussions* exhibits a low proportion of neutral sentiment, with high proportions of both positive and negative sentiment. This indicates that reviewer’s comments on the *Conclusions & Discussions* are likely to contain sentiment tendency.

### 4.2 Results of correlation analysis between review aspect sentiment and the duration of peer review across different disciplines

This section addresses *RQ1*. The study conducted a correlation analysis between review aspect sentiment and the duration of peer review for five first-level disciplines. To control for confounding factors affecting the duration of peer review, the study introduced article length, the number of authors, and the number of subjects as control variables. Pearson correlation coefficient was used to perform partial correlation analysis between sentiment scores of various review aspect clusters and the duration of peer review. The results are presented in Table 7 and Figure 3.

**Table 7 Results of partial correlation analysis between sentiment scores and the duration of peer review across different disciplines**

| **Review aspect clusters** | **Biological sciences** | **Earth and environmental sciences** | **Health sciences** | **Physical sciences** | **Scientific community and society** |
|---|---|---|---|---|---|
| Problem Definition /Idea**(PDI)** | -0.059** | -0.091** | -0.048** | -0.049** | -0.055 |
| Data/Datasets**(DAT)** | -0.083** | -0.099** | -0.085** | -0.107** | -0.087* |
| Tables & Figures **(TNF)** | -0.089** | -0.088** | -0.112** | -0.102** | -0.142** |
| Methodology & Model**(MET)** | -0.074** | -0.092** | -0.005 | -0.036** | -0.044 |
| Experiments & Operations**(EXP)** | -0.096** | -0.064* | -0.097** | -0.111** | -0.071 |
| Conclusions & Discussions**(CON)** | -0.100** | -0.099** | -0.104** | -0.108** | -0.026 |
| Index/Parameters **(IND)** | -0.062** | -0.059* | -0.082** | -0.085** | -0.044 |
| Evaluation & Result**(RES)** | -0.105** | -0.126** | -0.094** | -0.121** | -0.129** |
| Structure & Language**(SNL)** | -0.043** | -0.081** | -0.078** | -0.046** | -0.040 |
| Impact & Research Value**(IMP)** | -0.105** | -0.114** | -0.081** | -0.124** | -0.050 |
| Comparison & Correlation **(CMP)** | -0.073** | -0.068* | -0.058** | -0.073** | -0.079* |
| **Review** | -0.105** | -0.140** | -0.070** | -0.119** | -0.089* |

**Note**: * denotes significance at the 0.05 level (two-tailed), ** denotes significance at the 0.01 level (two-tailed).

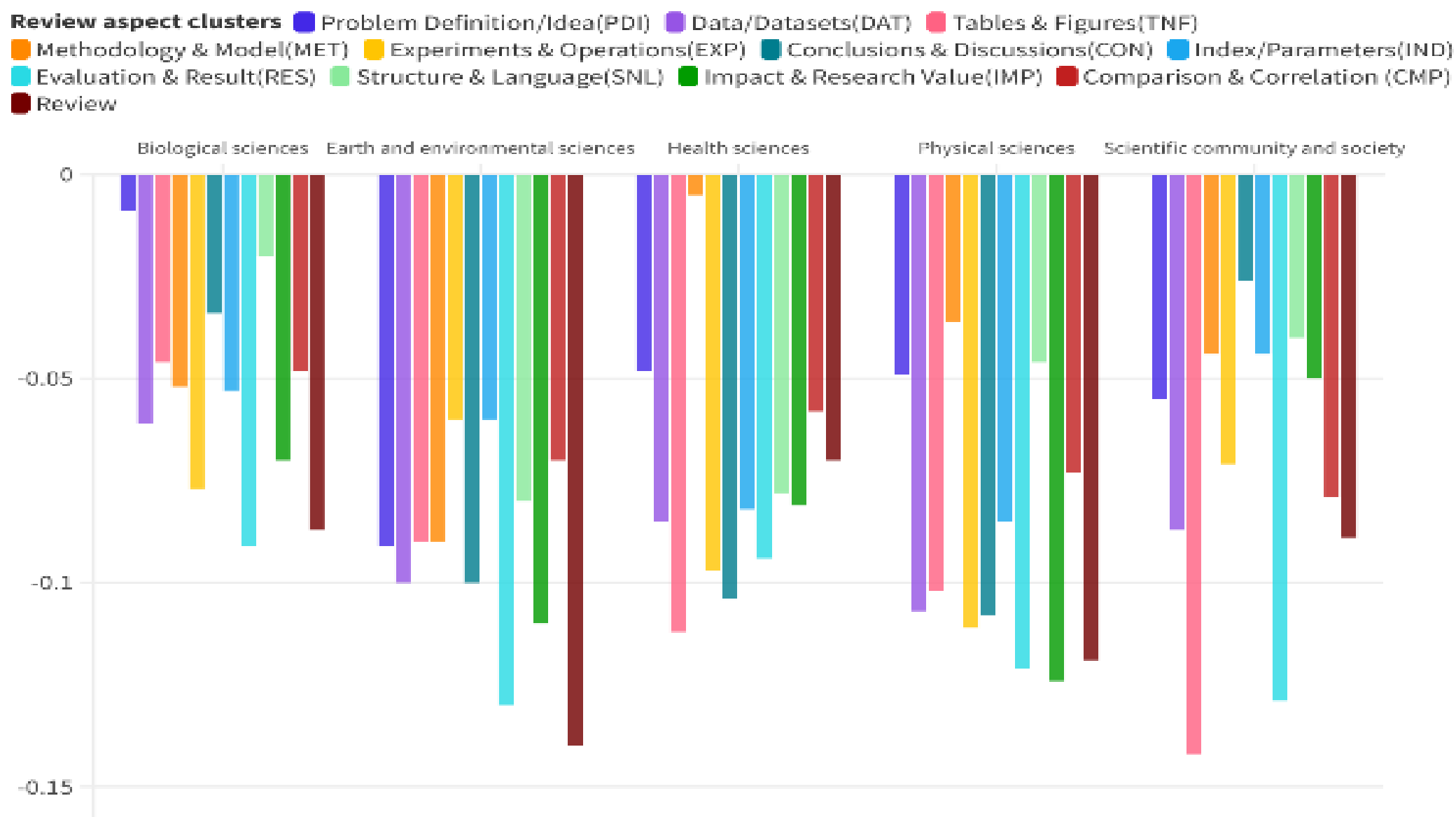

**Figure 3 Results of partial correlation analysis between sentiment scores and the duration of peer review across different disciplines**

From Table 7 and Figure 3, it is evident that there is a significant negative correlation between sentiment score of the overall review and the duration of peer review for each first-level discipline, but the correlation is very weak. This implies that papers with more positive review aspect sentiments are likely to have shorter the duration of peer reviews compared to those with more negative sentiments. Next, we will proceed to discuss and analyze the correlation between each review aspect cluster and the duration of peer review for each first-level discipline.

In Biological Sciences, there is a relatively stronger correlation between two review aspect clusters (*Evaluation & Result* and *Impact & Research Value*) and the duration of peer review. On the other hand, the correlation between review aspect cluster (*Structure & Language*) and the duration of peer review is relatively weaker. This suggests that reviewers in this discipline tend to pay more attention to the *Evaluation & Result* and *Impact & Research Value* of paper during the review process. The review and revision of these two aspects of paper will more likely take more time. Authors in this discipline should be particularly mindful of these aspects when writing and revising their papers; In Earth and environmental sciences, similar to Biological sciences, there is a relatively stronger correlation between two review aspect clusters (*Evaluation & Result* and *Impact & Research Value*) and the duration of peer review. However, what differs is that the correlation between two review aspect clusters (*Index/Parameters* and *Experiments & Operations*) and the duration of peer review is relatively weaker. This indicates that in this discipline, reviewers may not emphasize these two aspects of paper during the review process; In Health Sciences, there is a relatively stronger correlation between two review aspect clusters (*Tables & Figures* and *Conclusions & Discussions*) and the duration of peer review. However, the correlation between review aspect cluster (*Methodology & Model*) and the duration of peer review is extremely weak. This suggests that in this discipline, there is a relatively lower emphasis on the *Methodology & Model* of paper; In Physical Sciences, similar to the previous two disciplines, there is a relatively stronger correlation between two review aspect clusters (*Evaluation & Result* and *Impact & Research Value*) and the duration of peer review. On the other hand, the correlation between two review aspect clusters (*Methodology & Model* and *Structure & Language*) and the duration of peer review is relatively weaker; In Scientific Community and Society, there is a significant correlation between two review aspect clusters (*Tables & Figures* and *Evaluation & Result*) and the duration of peer review. However, the correlations between other review aspect clusters and the duration of peer review are relatively weak. It should be noted that the data volume in this discipline is relatively low, which may impact the results.

Across all disciplines, there's a relatively stronger correlation between two review aspect clusters (*Evaluation & Result* and *Impact & Research Value*) and the duration of peer review. Meanwhile, the correlation between review aspect cluster (*Structure & Language*) and the duration of peer review is the smallest among all review aspect clusters. This is partly because reviewers often focus on *Structure & Language* of paper in the later stages of the review process. Moreover, authors may not need to invest too much time in revising and optimizing *Structure & Language* of paper. On the contrary, the review and revision of the *Evaluation & Result* and *Impact & Research Value* of paper may more likely take much time, thus prolonging the Duration of Peer Review.

## 4.3 Results of correlation analysis between review aspect sentiment in different rounds and the duration of peer review

This section addresses *RQ2*. To explore the differences in the correlation between review aspect sentiment in different rounds and the duration of peer review in multiple peer review comments, this study extracted a corpus of peer review comments with three review rounds (due to the limited quantity of data with more than three review rounds, which accounted for only 2% of all corpus and was not suitable for analysis). The study then calculated sentiment scores of each review aspect cluster and the overall review for the first, second, and third round peer review comments. Pearson correlation coefficient was used to analyze the correlation between sentiment scores and the duration of peer review. The results are presented in Table 8 and Figure 4.

**Table 8 Results of correlation analysis between sentiment scores in different rounds and the duration of peer review**

| Review aspect clusters | First round | Second round | Third round |
|---|---|---|---|
| Problem Definition/Idea**(PDI)** | -0.009 | -0.061** | -0.046** |
| Data/Datasets**(DAT)** | -0.061** | -0.020 | -0.040** |
| Tables & Figures**(TNF)** | -0.046** | -0.112** | -0.063** |
| Methodology & Model**(MET)** | -0.052** | -0.047** | -0.015 |
| Experiments & Operations**(EXP)** | -0.077** | -0.037** | -0.038** |
| Conclusions & Discussions**(CON)** | -0.034** | -0.062** | -0.032** |
| Index/Parameters**(IND)** | -0.053** | -0.063** | -0.054** |
| Evaluation & Result**(RES)** | -0.091** | -0.068** | -0.010 |
| Structure & Language**(SNL)** | -0.020 | -0.080** | -0.053** |
| Impact & Research Value**(IMP)** | -0.070** | -0.016 | -0.023 |
| Comparison & Correlation **(CMP)** | -0.048** | -0.089** | -0.058** |
| **Review** | -0.087** | -0.063** | -0.042** |

**Note**: * denotes significance at the 0.05 level (two-tailed), ** denotes significance at the 0.01 level (two-tailed).

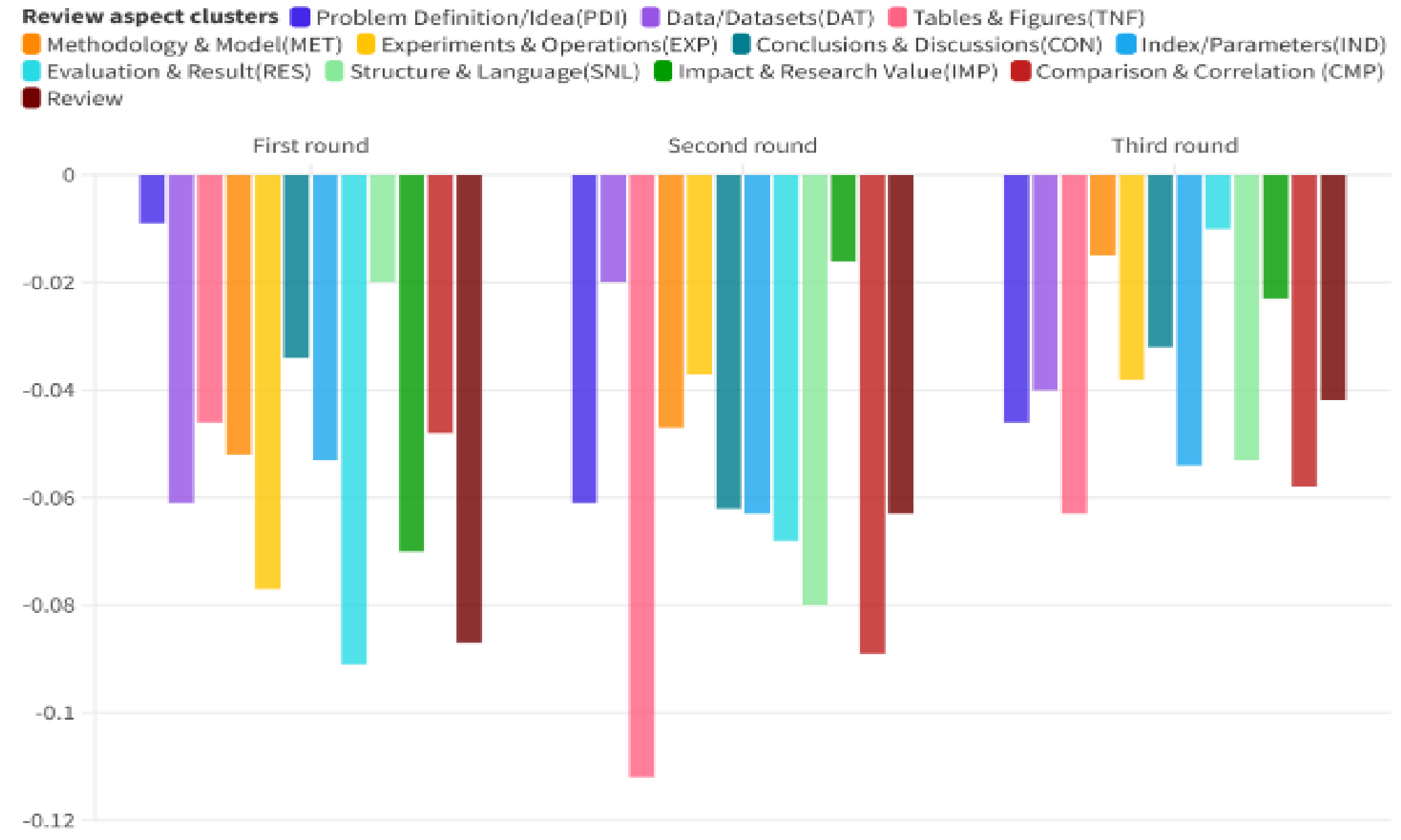

**Figure 4 Results of correlation analysis between sentiment scores in different rounds and the duration of peer review**

### (1) Correlation difference of the overall review across different rounds

From Table 8 and Figure 4, it can be observed that the correlation between sentiment scores of the overall review in the first round and the duration of peer review is relatively the strongest, followed by the second round, and the third round exhibits the weakest correlation. This indicates that in the peer review process, review aspect sentiment in the first round has the most significant impact on the duration of peer review. Therefore, the quality of the initial submission significantly influences the duration of peer review. The impact of review aspect sentiment in the second and third rounds on the duration of peer review decreases successively. This also reflects, from another perspective, that the time allocated to the first round review is the highest in the entire review process, followed by a decreasing proportion for each subsequent round. This is because the content of the first round peer review comments is typically the most comprehensive, with the content of peer review comments diminishing in subsequent rounds. For instance, in the first round peer review comments for the paper with DOI 10.1038/s41467-017-01815-7[3], there are 140 sentences containing review aspect sentiment, while the second round has 40 sentences, and the third round has 24 sentences.

### (2) Focus of reviewers in the first round

In the first round of peer review, the correlation between review aspect clusters (*Evaluation & Result, Experiments & Operations* and *Impact & Research Value*) and the duration of peer review is relatively the strongest. On the other hand, the correlation is relatively lower for review aspect clusters (*Problem Definition/Idea and Structure & Language*). This suggests that during the first round review process, reviewers tend to focus more on the content related to *Evaluation & Result, Experiments & Operations* and *Impact & Research Value*. In contrast, they pay less attention to *Problem Definition/Idea and Structure & Language* of paper. For example, in the first round peer review comments for the paper with

[3] https://static-content.springer.com/esm/art%3A10.1038%2Fs41467-017-01815-7/MediaObjects/41467_2017_1815_MOESM2_ESM.pdf

DOI 10.1038/s41467-017-00322-z[4], the reviewer provided the following comment regarding *Evaluation & Result* of paper: "*It is less accurate, difficult to analyze and results can be less reproducible.*" Therefore, when revising the paper in response to review comments in the first round, the author should prioritize addressing the reviewer's comments on *Evaluation & Result, Experiments & Operations a*nd *Impact & Research Value*.

**(3) Focus of reviewers in the second and third rounds**

In the second and third rounds of peer review, the correlation between review aspect clusters (*Tables & Figures, Comparison & Correlation and Structure & Language*) and the duration of peer review is relatively stronger. This correlation notably increases compared to the first round of peer review. Conversely, there is an opposite trend observed regarding aspect clusters like *Evaluation & Result, Impact & Research Value and Methodology & Model*. This suggests that during the second and third rounds of peer review, reviewers tend to shift their focus towards *Tables & Figures, Comparison & Correlation and Structure & Language* of paper, rather than emphasizing *Evaluation & Result, Impact & Research Value and Methodology & Model*. For instance, in the second round of peer review for the paper with the DOI 10.1038/s41467-017-00429-3 [5], the reviewer provided comment regarding *Comparison & Correlation* of paper: "*To actually make this point, the authors should strive to make this comparison between similar order arterioles from other organs, such as skeletal muscle.*" Similarly, in the third round of peer review for the paper with DOI 10.1038/s41467-017-01439-x[6], the reviewer provided comment regarding *Tables & Figures* of paper: "*However, I would suggest the modest change to the appearance of fig. 2e recommended below.*" Therefore, when revising the paper based on review comments in the second and third rounds, the authors should redirect their attention towards the reviewers' comments concerning *Tables & Figures, Comparison & Correlation and Structure & Language* of paper.

# 5. Discussion

This section will discuss the results and significance of this study, as well as elucidate the current limitations inherent in this study.

## 5.1 Implications

### 5.1.1 Theoretical implications

**(1) Connecting peer review comments with the peer review process.**

In recent years, with the gradual opening of peer review corpus, numerous scholars have conducted extensive research on peer review comments and the peer review process. However, there is limited research that explicitly connects peer review comments with the peer review process. Sarigöl et al. (2017) found that co-authorship between authors and journal editors shortens the duration of peer review. Liang et al. (2022) discovered that the stronger the novelty of a paper, the longer the processing time. Despite a considerable amount of research examining factors influencing the duration of peer review, there has been relatively little focus on review aspect sentiment factors inherent in peer review comments. This reflects a tendency to explore external factors influencing the duration of peer review while paying less attention to peer review comments themselves. This study establishes a connection between peer review comments and the peer review process, delving into review aspect sentiment in peer review comments and exploring their correlation with the duration of peer review. The results of this study indicate that *Evaluation & Result a*nd *Impact & Research Value* have a greater impact on the duration of peer review. Therefore, mining peer review comments related to these aspects is crucial for shortening the duration of peer review.

---

[4] https://static-content.springer.com/esm/art%3A10.1038%2Fs41467-017-00322-z/MediaObjects/41467_2017_322_MOESM1_ESM.pdf
[5] https://static-content.springer.com/esm/art%3A10.1038%2Fs41467-017-00429-3/MediaObjects/41467_2017_429_MOESM2_ESM.pdf
[6] https://static-content.springer.com/esm/art%3A10.1038%2Fs41467-017-00429-3/MediaObjects/41467_2017_429_MOESM2_ESM.pdf

**(2) Focusing on the characteristics of review round**

Peer review comments are often conducted through multiple rounds, forming texts resulting from iterative exchanges between reviewers and authors. Previous studies have commonly treated multi-round peer review comments as a whole. In contrast, this study divides the multi-round peer review comments into distinct review rounds and conducts an in-depth investigation into the characteristics of review round. We observed that reviewers exhibit different focus and sentiment distributions for paper across different review rounds. Therefore, exploring peer review comments from the perspective of different review rounds has significant value in research enrichment.

#### 5.1.2 Practical implications

**(1) Providing a reference for authors to revise and optimize their papers**

Peer review is a crucial process that academic papers must go through for publication and is of great importance to paper submitters. However, many authors often struggle to understand the focus of reviewers during the peer review process. It can lead to less-than-satisfactory revisions and optimizations of paper, making it challenging to pass the final acceptance decision. This study employs aspect sentiment analysis to uncover the specific aspects that reviewers in different disciplines focus on. Additionally, it reveals that the emphasis of reviewers varies across different rounds of the peer review process. Authors can use this information as a reference when revising and optimizing their papers, tailoring revisions to address the focus of reviewers across different review rounds.

**(2) Providing a reference for improving the efficiency of peer review**

In recent years, due to the rapid increase in the number of submitted papers, the peer review system has become overwhelmed, making it a particularly important issue to improve the efficiency of peer review. This study explores the correlation between review aspect sentiment and the duration of peer review to identify aspects that have a greater impact on the duration of peer review. It provides assistance from the perspectives of authors, reviewers, and editors/chairs to enhance the efficiency of peer review. Regarding aspects that significantly affect the duration of peer review, authors need to pay special attention to ensuring quality during writing and optimization; Reviewers should provide specific and effective suggestions for these aspects to assist authors in revising and optimizing. Editors/chairs can make final decisions by closely considering the reviews and optimization outcomes related to these aspects.

## 5.2 Limitations

This study is based on peer review comments from *Nature Communications*, investigating the correlation between review aspect sentiment and the duration of peer review in multi-round peer review. The study has yielded some valuable findings, but there are currently several limitations, mainly including the following aspects:

**(1) Data completeness**
The data utilized in this study are sourced from *Nature Communications*, which only publicly discloses peer review comments for some of its accepted papers, lacking access to peer review comments for its rejected papers. Furthermore, the data collected for this study only extends until the year 2020 and does not include the most recent publications."

**(2) Sentiment classification performance**
Currently, supervised deep learning models are the mainstream methods used for automatic sentiment classification. However, their performance is highly dependent on the quality and scale of the manually annotated training corpus. The model that achieved the highest performance on the annotated corpus used in this study was proposed in 2019, and there have been continuous advancements with new and improved models introduced in recent years."

**(3) Comprehensive of control variables**
Factors influencing the duration of peer review are highly complex and may be related to various aspects such as the subjects of paper, reviewers, journals/conferences, and authors themselves. This study only considered three factors, namely article length, the number of authors, and the number of subjects, as control variables. It did not comprehensively consider various factors that could potentially impact the duration of peer review.

# 6. Conclusion and Future work

The main conclusions of this study are as follows: Firstly, there is a significant negative correlation between review aspect sentiment and the duration of peer review, but the correlation is very weak. In other words, papers with more positive review aspect sentiments are likely to have shorter the duration of peer reviews compared to those with more negative sentiments. Secondly, there is a relatively stronger correlation between two review aspect clusters (*Evaluation & Result* and *Impact & Research Value*) and the duration of peer review. Meanwhile, the correlation between review aspect cluster (*Structure & Language*) and the duration of peer review is the smallest among all review aspect clusters. Therefore, when authors are revising and optimizing their papers, it is crucial to focus on *Evaluation & Result* and *Impact & Research Value* of paper. Finally, in the first round of peer review, reviewers tend to focus more on the content related to *Evaluation & Result, Experiments & Operations a*nd *Impact & Research Value*. In the second and third rounds of peer review, reviewers tend to shift their focus towards *Tables & Figures, Comparison & Correlation and Structure & Language* of paper. Hence, when authors are revising and optimizing their papers in different review rounds, they should make targeted revisions based on the specific aspects that reviewers emphasize in different review rounds.

In the future, this study aims to address the current limitations by implementing the following improvements: Firstly, we will utilize more journals/conferences that publicly disclose peer review comments for both accepted and rejected papers as data sources. The data collection will also be updated to include the most recent year. Secondly, we will improve both the quality and scale of manually annotated training data. Simultaneously, various recently proposed sentiment classification models will be explored using our manually annotated data for training and testing. The goal is to achieve higher classification performance for the sentiment classification task on all corpus. Finally, we will conduct further research in the peer review field, considering various factors comprehensively, including the subjects of paper, reviewers, journals/conferences, and authors themselves. Efforts will be made to quantitatively measure these factors and use them as control variables for analysis and research.

## Acknowledgment

## References

Benos, D. J., Bashari, E., Chaves, J. M., Gaggar, A., Kapoor, N., LaFrance, M., ... & Zotov, A. (2007). The ups and downs of peer review. Advances in physiology education, 31(2), 145-152.

Bharti, P. K., Ghosal, T., Agarwal, M., & Ekbal, A. (2023). PEERRec: An AI-based approach to automatically generate recommendations and predict decisions in peer review. International Journal on Digital Libraries, 1-18. https://doi.org/10.1007/s00799-023-00375-0

Bhatia, C., Pradhan, T., & Pal, S. (2020, July). Metagen: An academic meta-review generation system. In Proceedings of the 43rd International ACM SIGIR Conference on Research and Development in Information Retrieval, Virtual Event, China, pp. 1653-1656.

Björk B, Solomon D. (2013). The publishing delay in scholarly peer-reviewed journals. Journal of Informetrics, 7:914-923.

Bornmann L, Daniel HD. (2010). How long is the peer review process for journal manuscripts? A case study on Angewandte Chemie international edition. CHIMIA International Journal for Chemistry, 64(1), 72-77.

Chakraborty, S., Goyal, P., & Mukherjee, A. (2020, August). Aspect-based sentiment analysis of scientific reviews. In Proceedings of the ACM/IEEE Joint Conference on Digital Libraries in 2020, Wuhan, China, pp. 207-216.

Chong, S.W. (2021), Improving peer-review by developing reviewers' feedback literacy. Learned Publishing, 34: 461-467.

Day, A. (2022). Exploratory analysis of text duplication in peer-review reveals peer-review fraud and paper mills. Scientometrics, 127(10), 5965-5987.

Devlin, J., Chang, M. W., Lee, K., & Toutanova, K. (2019, June). Bert: Pre-training of deep bidirectional transformers for language understanding. In: Proceedings of the 2019 Conference of the North American Chapter of the Association for Computational Linguistics: Human Language Technologies, Minneapolis, United States, pp. 4171-4186.

Frey, B. J., & Dueck, D. (2007). Clustering by passing messages between data points. science, 315(5814), 972-976.

Fromm, M., Faerman, E., Berrendorf, M., Bhargava, S., Qi, R., Zhang, Y., Dennert, L., Selle, S., Mao,

Y., & Seidl, T. (2021, May). Argument Mining Driven Analysis of Peer-Reviews. *Proceedings of the AAAI Conference on Artificial Intelligence*, *35*(6), 4758-4766.

Ghosal, T., Kumar, S., Bharti, P. K., & Ekbal, A. (2022). Peer review analyze: A novel benchmark resource for computational analysis of peer reviews. *Plos one*, *17*(1), e0259238.

Ghosal, T., Verma, R., Ekbal, A., & Bhattacharyya, P. (2019a, June). A sentiment augmented deep architecture to predict peer review outcomes. In 2019 ACM/IEEE Joint Conference on Digital Libraries (JCDL), Champaign, USA, pp. 414-415.

Ghosal, T., Verma, R., Ekbal, A., & Bhattacharyya, P. (2019b, July). DeepSentiPeer: Harnessing sentiment in review texts to recommend peer review decisions. In Proceedings of the 57th Annual Meeting of the Association for Computational Linguistics, Florence, Italy, pp. 1120-1130.

Han, R., Zhou, H., Zhong, J. & Zhang, C. (2022, October). Characterizing Peer Review Comments of Academic Articles in Multiple Rounds. Proceedings of the 85th Annual Meeting of the Association for Information Science and Technology, Pittsburgh, USA, pp.89-99.

Han, R., Zhou, H., Zhong, J., Qin, C. & Zhang, C.(2023, July). Aspect Sentiment Distribution and Change in Multiple Rounds of Peer Review Comments. In: Proceedings of the 19th International Conference on Scientometrics and Informetrics (ISSI 2023), Bloomington, USA.

He, Y., Tian, K., & Xu, X. (2023). A validation study on the factors affecting the practice modes of open peer review. Scientometrics, 128(1), 587-607.

Hecht, B., Wilcox, L., Bigham, J. P., Schöning, J., Hoque, E., Ernst, J., ... & Wu, C. (2021). It's time to do something: Mitigating the negative impacts of computing through a change to the peer review process. arXiv preprint arXiv:2112.09544.

Helmer, M., Schottdorf, M., Neef, A., & Battaglia, D. (2017). Gender bias in scholarly peer review. Elife, 6, e21718.

Hu, M., & Liu, B. (2004, August). Mining and summarizing customer reviews. In Proceedings of the tenth ACM SIGKDD international conference on Knowledge discovery and data mining, New York, USA, pp. 168-177.

Hua, X., Nikolov, M., Badugu, N., & Wang, L. (2019). Argument mining for understanding peer reviews. arXiv preprint arXiv:1903.10104.

Huang, B., Ou, Y., & Carley, K. M. (2018, July). Aspect level sentiment classification with attention-over-attention neural networks. In: Proceedings of the 11th International Conference on Social Computing: Behavioral-cultural Modeling, and Prediction Conference and Behavior Representation in Modeling and Simulation (SBP-BRIMS 2018), Washington, United States, pp. 197-206.

Huisman, J., Smits, J. Duration and quality of the peer review process: the author's perspective. *Scientometrics* **113**, 633–650 (2017).

Kang, D., Ammar, W., Dalvi, B., Van Zuylen, M., Kohlmeier, S., Hovy, E., & Schwartz, R. (2018, June). A dataset of peer reviews (peerread): Collection, insights and nlp applications. In: Proceedings of the 2018 Conference of the North American Chapter of the Association for Computational Linguistics: Human Language Technologies, New Orleans, United states, pp. 1647-1661.

Kousha, K., & Thelwall, M. (2023). Artificial intelligence to support publishing and peer review: A summary and review. Learned Publishing. https://doi.org/10.1002/leap.1570

Lee, C. J., Sugimoto, C. R., Zhang, G., & Cronin, B. (2013). Bias in peer review. Journal of the American Society for information Science and Technology, 64(1), 2-17.

Liang, Z., Mao, J., & Li, G. (2022). Bias against scientific novelty: A prepublication perspective. Journal of the Association for Information Science and Technology, 74(1), 99-114.

Liu, B. (2015). Sentiment Analysis: Mining Opinions, Sentiments, and Emotions. Cambridge: Cambridge University Press.

Liu, Q., Zhang, H., Zeng, Y., Huang, Z., & Wu, Z. (2018, April). Content attention model for aspect based sentiment analysis. In Proceedings of the 2018 world wide web conference, Lyon, France, pp. 1023-1032.

Luo, J., Feliciani, T., Reinhart, M., Hartstein, J., Das, V., Alabi, O., & Shankar, K. (2021). Analysing sentiments in peer review reports: evidence from two science funding agencies. Quantitative Science Studies, 2(4), 1271-1295.

Lyman R.L. (2013). A three-decade history of the duration of peer review. Journal of Scholarly Publishing, 44(3), 211-220.

Ma, D., Li, S., Zhang, X., & Wang, H. (2017). Interactive attention networks for aspect-level sentiment classification. arXiv preprint arXiv:1709.00893.

McHugh M. L. (2012). Interrater reliability: the kappa statistic. Biochemia medica, 22(3), 276–282.

Meng M., Wang Y., Zhang C. (2021, July). Building Multi-level Aspects of Peer Reviews for Academic Articles. In: Proceedings of the 18th International Conference on Scientometrics and Informetrics (ISSI 2021), Online, pp. 1519-1520.

Mrowinski, M. J., Fronczak, A., Fronczak, P., Nedic, O., & Ausloos, M. (2016). The duration of peer review in peer review: Quantitative analysis and modelling of editorial workflows. Scientometrics, 107(1), 271-286.

Ni, J., Zhao, Z., Shao, Y., Liu, S., Li, W., Zhuang, Y., ... & Li, J. (2021). The influence of opening up peer review on the citations of journal articles. Scientometrics, 126(12), 9393-9404.

Price, S., & Flach, P. A. (2017). Computational support for academic peer review: A perspective from artificial intelligence. Communications of the ACM, 60(3), 70-79.

Qin, C., & Zhang, C. (2021, July). Exploring the Distribution of Referees' Comments in IMRaD Structure of Academic Articles. In: Proceedings of the 18th International Conference on Scientometrics and Informetrics (ISSI 2021), Online, 2021.

Qin, C. and Zhang, C. (2023), "Which structure of academic articles do referees pay more attention to?: perspective of peer review and full-text of academic articles", Aslib Journal of Information Management, 75(5), 884-916.

Qiu, G., Liu, B., Bu, J., & Chen, C. (2011). Opinion word expansion and target extraction through double propagation. Computational linguistics, 37(1), 9-27.

Ramachandran, L., & Gehringer, E. F. (2011, July). Automated assessment of review quality using latent semantic analysis. In 2011 IEEE 11th International Conference on Advanced Learning Technologies, Zhengzhou, China, pp. 136-138.

Sabaj, O.; Valderrama, J.O.; González, C. & Pina-Stranger, A. (2015). Relationship between the Duration of Peer- Review, Publication Decision, and Agreement among Reviewers in three Chilean Journals. European Science Editing, 41 (4), 87-90.

Sabaj, O.; González, C. & Pina-Stranger, A. (2016). What we Still don't Know about Peer Review. Journal of Scholarly Publishing, 47 (2), 180-212.

Sarigöl, E., Garcia, D., Scholtes, I., & Schweitzer, F. (2017). Quantifying the effect of editor–author relations on manuscript handling times. Scientometrics, 113(1), 609-631.

Segarra-Saavedra, J., Hidalgo-Marí, T. and Tur-Viñes, V. (2023), Editorial time management: Peer review dates and other key dates of Spanish Communication journals. Learned Publishing.

Severin, A. and Chataway, J. (2021), Purposes of peer review: A qualitative study of stakeholder expectations and perceptions. Learned Publishing, 34: 144-155.

Song, Y., Wang, J., Jiang, T., Liu, Z., & Rao, Y. (2019, September). Targeted sentiment classification with attentional encoder network. In Artificial Neural Networks and Machine Learning–ICANN 2019: Text and Time Series: 28th International Conference on Artificial Neural Networks, Munich, Germany, pp. 93-103.

Squazzoni, F., Bravo, G., Grimaldo, F., García-Costa, D., Farjam, M., & Mehmani, B. (2021). Gender gap in journal submissions and peer review during the first wave of the COVID-19 pandemic. A study on 2329 Elsevier journals. PloS one, 16(10), e0257919.

Tang, D., Qin, B., & Liu, T. (2016). Aspect level sentiment classification with deep memory network. arXiv preprint arXiv:1605.08900.

Tennant, J. P. (2018). The State of the Art in Peer Review. Fems Microbiology Letters, 365(19), 1-10.

Thelwall, M., Papas, E. R., Nyakoojo, Z., Allen, L., & Weigert, V. (2020). Automatically detecting open academic review praise and criticism. Online Information Review, 44(5), 1057-1076.

Thelwall, M., & Kousha, K. (2023). Technology assisted research assessment: algorithmic bias and transparency issues. Aslib Journal of Information Management, (ahead-of-print).

Varas Espinoza, G., Sabaj Meruane, O. & Pina-Stranger, A. (2020). Feedback quality according to the type of referees in the peer review process of scientific articles. Revista Hipatia, 2, 37-60.

Vincent-Lamarre, P., & Larivière, V. (2021). Textual analysis of artificial intelligence manuscripts reveals features associated with peer review outcome. Quantitative Science Studies, 2(2),662-677.

Waltman, L., Kaltenbrunner, W., Pinfield, S. and Woods, H.B. (2023), How to improve scientific peer review: Four schools of thought. Learned Publishing, 36: 334-347.

Wang, Y., Huang, M., Zhu, X., & Zhao, L. (2016, November). Attention-based LSTM for aspect-level sentiment classification. In Proceedings of the 2016 conference on empirical methods in natural language processing, Austin, United States, pp. 606-615.

Wang, K., & Wan, X. (2018, June). Sentiment analysis of peer review texts for scholarly papers. In The 41st International ACM SIGIR Conference on Research & Development in Information Retrieval, Ann Arbor, United states, pp. 175-184.

Wang, Q., Zeng, Q., Huang, L., Knight, K., Ji, H., & Rajani, N. F. (2020). ReviewRobot: Explainable paper review generation based on knowledge synthesis. arXiv preprint arXiv:2010.06119.

Wu, W., Xi, H. & Zhang, C. (2024). Are the confidence scores of reviewers consistent with the review content? Evidence from top conference proceedings in AI. Scientometrics, 129(7): 4109–4135 https://doi.org/10.1007/s11192-024-05070-8.

Zeng, B., Yang, H., Xu, R., Zhou, W., & Han, X. (2019). Lcf: A local context focus mechanism for aspect-based sentiment classification. Applied Sciences, 9(16), 3389.

Zhang, G., Wang, L., Xie, W., Shang, F., Xia, X., Jiang, C. and Wang, X. (2022), "“This article is interesting, however”: exploring the language use in the peer review comment of articles published in the BMJ", Aslib Journal of Information Management, 74(3), 399-416.